\documentclass[letterpaper, 10 pt, conference]{ieeeconf}  

\IEEEoverridecommandlockouts                              
\usepackage{graphicx} 
\usepackage{amsmath} 
\usepackage{hyperref}
\usepackage{algorithm}
\usepackage[noend]{algpseudocode}

\title{\LARGE \bf
Real Time Collision Detection and Identification\\for Robotic Manipulators
}

\author{Elena Galbally and Mikael Jorda
}

\begin{document}

\maketitle
\thispagestyle{empty}
\pagestyle{empty}

\begin{abstract}
The majority of everyday tasks involve interacting with unstructured environments. This implies that, in order for robots to be truly useful they must be able to handle contacts. This paper explores how a particle filter can be used to localize a contact point and estimate the external force. We demonstrate the capability of the particle filter on a simulated 4DoF planar robotic arm, and compare it to a well-established analytical approach.

\end{abstract}

\section{INTRODUCTION}

For several decades, robots have been able to reliably follow precise trajectories making them ideal tools for assembly lines. However, in 2017, we still do not see any complex robot capable of manipulating objects in unstructured environments such as homes or offices. Robots remain confined to industrial settings. One of the main factors preventing robots from conquering other types of environments is their inability to robustly handle single or multi contact collisions. In fact, as seen for example during the DARPA Robotics Challenge in 2015, contacts lead to robot failures and safety concerns.

In order to be able to handle contacts, robots must first gain the ability to detect and identify them. Several solutions have been proposed in the past years. Some approaches involve using tactile skins to detect the contacts \cite{Dahiya2013}. This solution is ideal in theory, however, there are still no commercially available tactile skins that are reliable, and affordable. Therefore, this approach is usually limited to robotic hands which is not enough to handle situations where the arm of the robot touches the environment. Other proposed solutions use Force/Torque sensors pre-positioned on the robot (usually at the wrist or base) \cite{Lu2005} \cite{Yamada2009}, but once again this approach is limited by the locations of the force sensors selected during the design phase.

A different approach using only proprioceptive sensors was introduced in 1989 \cite{Takakura1989}. The idea was to compare the applied joint torques to the sensed torques. Disturbance torques due to contact would make these two differ. Using this approach, the sensed free space torques are simply obtained from the robot's dynamic equation. This idea has been refined to get more robust and easy to compute observers \cite{DeLuca2005} (the main advantage being that they do not need acceleration measurements of the joints anymore). Additionally, it has been extended to mobile manipulators \cite{Kim2016}, flexible joint robots \cite{Lee2016}, and humanoids \cite{Vorndamme2017}. Recently, people have also addressed the problem using particle filters \cite{Manuelli2016} with promising results.

In this paper, we propose to compare the analytical approach described in \cite{Haddadin2017} and the particle filter approach in \cite{Manuelli2016} for the problem of contact detection and localization in the context of controlling a multi DoF planar robot to perform a certain task while in contact with the environment.

\section{PROBLEM FORMULATION}
\begin{figure}[h]
\centering
\includegraphics[width = 0.9\columnwidth]{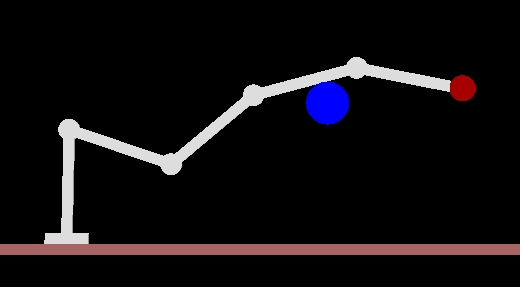}
\caption{Planar robot performing a task at its end effector (red circle) while bracing on the environment (blue circle).}
\label{setup}
\end{figure}

For a robot performing a task in free space or in contact, it can be advantageous to brace on the environment in order to save power and to reduce the uncertainties in the movement. However, once in contact with the environment, the dynamics of the robot change and we need to account for this variation in our control law. An example is shown fig \ref{setup}, where we would like to know both the exact contact point location between the blue shape and the robot, and the value of the contact force. This procedure is comprised of three steps :
\begin{itemize}
\item Collision detection - consists on detecting that a collision occurred and identifying the related disturbance torques. 
\item Collision isolation - involves finding the exact location of the collision.
\item Collision identification - requires quantifying the force at the collision.
\end{itemize}

\subsection{MOMENTUM OBSERVER FOR COLLISION DETECTION}
Drawing inspiration from \cite{Haddadin2017}, we will use a momentum based observer for the collision detection step. Let us review this approach here. The dynamics equation of an articulated rigid body manipulator is:

\begin{align}
M(q)\ddot{q} + C(q,\dot{q})\dot{q} + g(q) = \tau_m + \tau_c
\end{align}

where $M(q)$ is the robot mass matrix, $C(q,\dot{q})\dot{q}$ is the centrifugal and Coriolis vector, factorized with the matrix $C$ of Christoffel's symbols, $g(q)$ is the vector of joint torques, $\tau_c$ is the vector of contact torques (torques on the joints due to contacts with the environment), and $\tau_m$ is the vector of motor torques. In the following, we will systematically omit the dependencies on $q$ when writing these terms so $M(q) = M$, $C(q,\dot{q}) = C$, and $g(q) = g$. The generalized momentum of the robot is given by:

\begin{align}
p = M\dot{q}
\end{align}

Therefore, 
\begin{align}
\dot{p} = \dot{M}\dot{q} + M\ddot{q}.
\end{align}

Combining this expression with the dynamics equation yields:

\begin{align}
\dot{p} = \tau_m + \tau_c - g - C\dot{q} + \dot{M}\dot{q}
\end{align}

A basic property of the robot mass matrix is the skew-symmetry of $\dot{M}-2C$ which is equivalent to the equation $\dot{M} = C + C^T$. Hence, we have:

\begin{align}
\dot{p} = \tau_m + \tau_c - g + C^T\dot{q}
\end{align}

Let us define the residual vector $\gamma$ as:
\begin{align}
\gamma = K[p + \int(g - C^T\dot{q} - \tau_m - \gamma)dt]
\end{align}

From this definition, it follows that: 
\begin{align}
\dot{\gamma} = K(\tau_c - \gamma).
\end{align}

With diagonal $K > 0$, this is a stable, linear, decoupled, first-order estimation of the contact torques $\tau_c$, and if $K$ is sufficiently large, we can assume $\gamma \approx \tau_c$.

When the robot moves in free space, all the coefficients of $\gamma$ will be close to zero. When a collision occurs, all the coefficients of $\gamma$ corresponding to the joints that come before the collision point will become non-zero. Computing $\gamma$ is a good way to get three  key pieces of information :
\begin{itemize}
\item Whether a collision is happening
\item At which link the collision is happening
\item What are the corresponding torques at the joints
\end{itemize}

Now, we will present two different ways of performing the collision isolation and identification that make use of this momentum observer. 

\subsection{COLLISION IDENTIFICATION AND ISOLATION : \\ ANALYTICAL APPROACH}
\begin{figure}[h]
\centering
\includegraphics[width = 0.9\columnwidth]{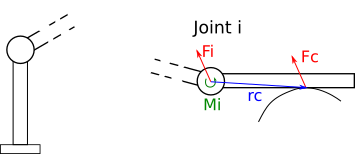}
\caption{Schematic explaining the analytical solution. We can resolve the contact torques $\gamma$ into force and moment at the joint frame, and then use these values and the link geometry to get the contact position.}
\label{fig::analytical}
\end{figure}

We assume a single contact point. This implies that there is a contact force $F_c$, but no moment. Let $J_c$ be the Jacobian matrix at the contact point, then we know $\gamma = J_c^TF_c$. The problem is that we don't know neither $J_c$ (it depends on the exact contact point) nor $F_c$. Nevertheless, we can compute the equivalent force and moment at any point where we do know the value of jacobian, such as the joint (see fig \ref{fig::analytical}). If we choose the base link, then there is a force $F_i$ and a moment $M_i$ such that:

\begin{align}
\gamma = J_i^T
\begin{bmatrix}
F_i \\ M_i
\end{bmatrix}
\label{eq::Jtf_link}
\end{align}

and we have the relations
\begin{align}
F_i &= F_c \\
M_i &= r_c \times F_c
\end{align}

If we can invert equation \ref{eq::Jtf_link} to get $F_i$ and $M_i$, then we will know both the force applied at the contact point and the line of action (which can be extracted from $M_i$). Knowing the line of action and the geometry of the link, we get two possible contact points. Finally, the constraint that the contact force can only push the robot away from the obstacle allows us to chose between these two points. In general, this is only possible if $J_i^T$ has more than 3 rows, which means that we need at least 3 proprioceptive sensors between the contact point and the base of the robot.

This method has the advantage of being easy to implement and fast to compute. However, it is difficult to generalize to multiple contacts, and it cannot identify successfully contacts on the first links of the manipulator.

\subsection{COLLISION IDENTIFICATION AND ISOLATION : \\PARTICLE FILTER APPROACH}
Another approach is to use a particle filter to locate the contact and estimate the force at that point. In \cite{Manuelli2016}, such a filter is used to obtain the contact location only. However, as described below, we can easily adapt this method to also get an estimate of the contact force. Once again, we start from the momentum observer $\gamma$. The idea here is to consider the possible contact locations as our state space, and to use a particle filter without rejection to converge to the true state. 

\paragraph{\textit{Measurement Model}}
We want to know how well a particle explains the observed measurement $\gamma$. So we want to know, for each particle location $r_i$, if there is a contact force $F_c$ such that $\gamma = J_i^TF_c$ where $J_i$ is the Jacobian at $r_i$. Thus, we have the following minimization problem :

\begin{align}
\min_{F_c}\|\gamma-J_i^TF_c\|_2^2 \\
\textrm{Subject to } F_c \in S(i)
\end{align}

Where $S(i)$ is the half plane that creates forces that push the robot away from the environment at particle location $i$. This is a quadratic program (QP) with a linear inequality constraint, and there exist many solvers for these sort of problems. The argument of the minimization is the contact force we are looking for. Finally, our measurement model is:

\begin{align}
P(\gamma|r_i) \propto \exp(-\dfrac{1}{2}QP(\gamma|r_i))
\end{align}

We must normalize the expression above to get the weights. 

\paragraph{\textit{Motion Model}}
We will use a simple model where the particles move on the surface of the link with some noise that follows a Gaussian distribution.

\begin{align}
r_i^{t+1} = r_i^t + \alpha d
\end{align}

Where $d$ is the vector that follows the boundary of the link on which $r_i^{t}$ is, and $\alpha \sim N(0,\sigma^2)$ is Gaussian noise with mean zero and variance $\sigma^2$. 

\begin{figure}
\centering
\includegraphics[width=\columnwidth]{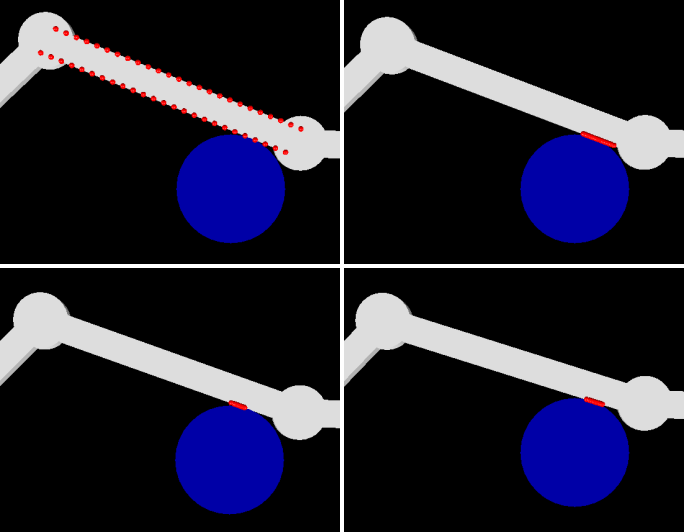}
\caption{Example of the particle set (in red) from the prior uniform distribution to the estimate of the actual contact point in 4 consecutive timesteps.}
\label{pfilter}
\end{figure}

The algorithm is outlined below and illustrated in (fig \ref{pfilter}).

\makeatletter
\def\BState{\State\hskip-\ALG@thistlm}
\makeatother

\begin{algorithm}
\caption{Single Contact Particle Filter}\label{particleFilter}
\begin{algorithmic}[1]
\BState \emph{Contact detection}:
\If {$\gamma_t < \epsilon_{mu} $} 
\State $X_t = \emptyset$
\State \Return $X(t)$
\EndIf\textbf{end if}
\State
\BState \emph{Sample particles}:
\If {$X_{t-1} = \emptyset$}
\State $X_{t}' = X_{init}$ \Else 
\State $X_{t}' = MotionModel(X_{t-1})$
\EndIf\textbf{end if}
\State
\BState \emph{Resample according to weights}:
\For {$r_i^{t}$ in $X_{t}'$}
\State $w_i^{t} = P(\gamma|r_i^{t})$
\EndFor\textbf{end for}
\State $X_t = Resample(X_{t}')$
\State
\State \Return $X_t$
\end{algorithmic}
\end{algorithm}

Where $\epsilon_{mu}$ is the threshold force value above which we consider that there has been a collision, $\gamma_t$ is our momentum observer measurement, and $X_{init}$ is a set of uniformly distributed samples on the surface of the link (in this case we used 50 particles). As our resampling strategy we used a multimodal approach in which you choose a particle with a probability proportional to its weight by generating a random number \cite{resampling}.

\section{EXPERIMENTS IN SIMULATION}
We compare the performance of the two previous approaches in simulation, using the simulation environment SAI2 developed in the Stanford Robotics Lab. We will test two scenarios in which the robot is controlled to perform a task at the end effector while the fourth link of the robot is in contact with the environment. 

\begin{figure}
\centering
\includegraphics[width=\columnwidth]{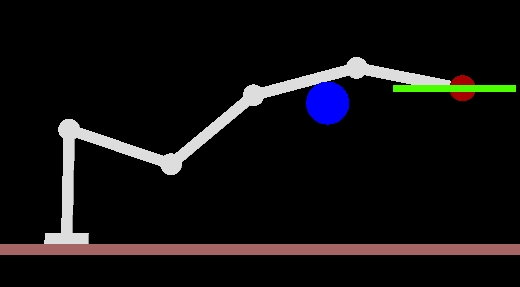}
\caption{In green : the trajectory that the end effector follows in the second scenario}
\label{fig::traj2d}
\end{figure}

In the first scenario, the task is simply to hold the end effector position. In the second scenario, the task consists on tracking a trajectory (fig \ref{fig::traj2d}). To perform these tasks we use the operational space formulation \cite{Khatib1987}, which allows us to directly control the end effector of the robot with a PD controller. For both tasks, the simulation starts with the robot in free space and then falls after a certain time and collides with the blue object (see the attached video).

In all of our experiments, we implemented a Butterworth filter of order 2 at 15Hz for both the forces and positions. This allows us to filter out noise and obtain smoother plots.

\section{RESULTS AND ANALYSIS}
\subsection{Task performance figures}  
We used continuous lines to represent the real quantities, while the estimated quantities are dashed lines. 

\begin{figure}
\centering
\includegraphics[width=\columnwidth]{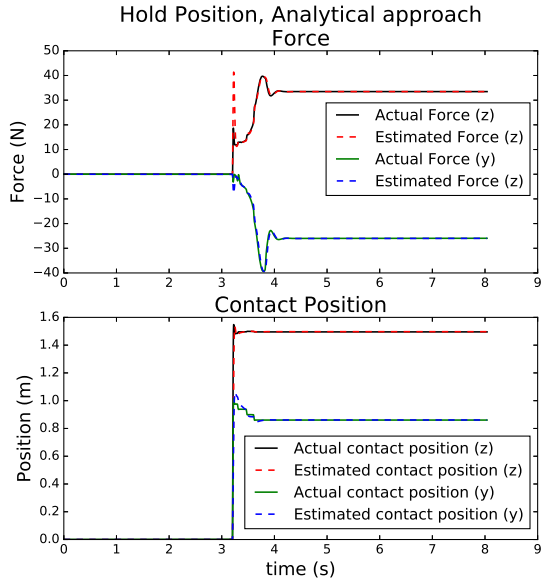}
\caption{Estimated force and real force in the YZ plane for the holding position scenario using the analytical method}
\label{hpos_analytical}
\end{figure}

\begin{figure}
\centering
\includegraphics[width=\columnwidth]{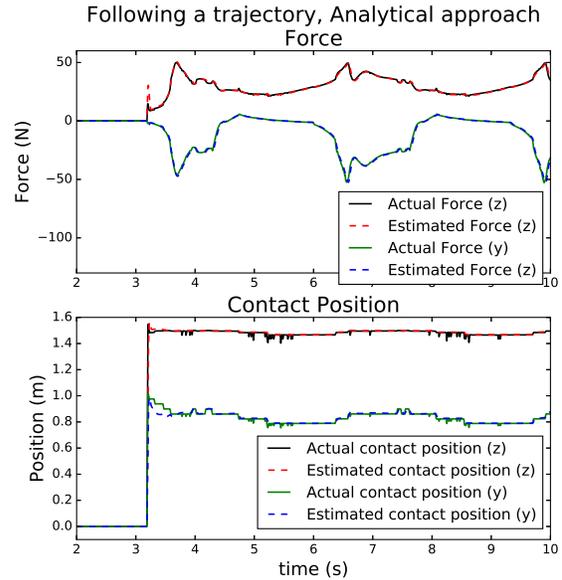}
\caption{Estimated force and real force in the YZ plane for the following trajectory scenario using the analytical method}
\label{ftraj_analytical}
\end{figure}

\begin{figure}
\centering
\includegraphics[width=\columnwidth]{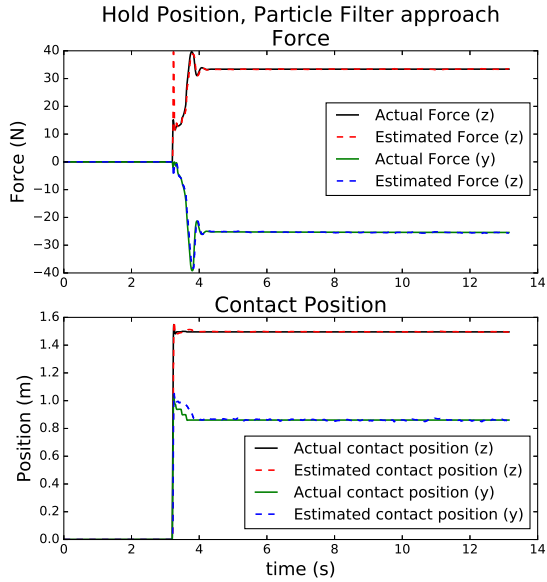}
\caption{Estimated force and real force in the YZ plane for the holding position scenario using the particle filter method}
\label{hpos_pfilter}
\end{figure}

\begin{figure}
\centering
\includegraphics[width=\columnwidth]{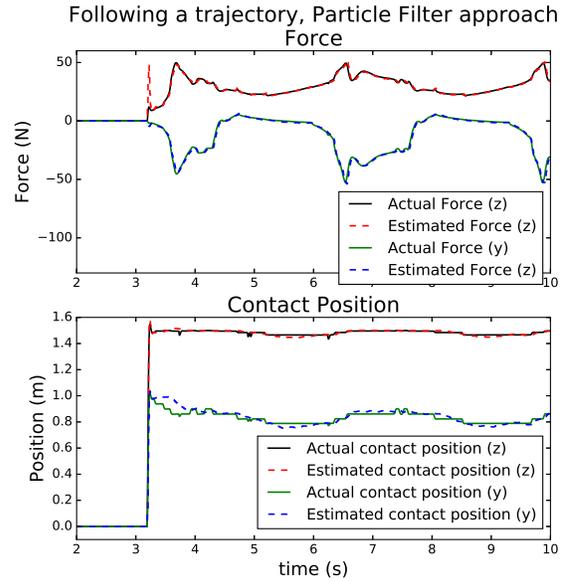}
\caption{Estimated force and real force in the YZ plane for the following trajectory scenario using the particle filter  method}
\label{ftraj_pfilter}
\end{figure}

As we can see in the plots for both methods and tasks, there is an initial peak in the estimated force values at the moment of the collision. This is expected because the impact causes a discontinuity in the dynamics that the momentum observer is not able to instantaneously account for. Other than this, the estimates seem really close to the true values in both cases, with the force estimate being more accurate than the position (the errors are less than 0.5 N and less than 1 cm in the position).

\subsection{Localization and force estimation} 
For both tasks, as illustrated in the plots, the methods used are able to accurately locate the contact point and provide a reasonable estimate for $F_c$. Hence, it is possible to say that they perform equally well in this planar example when the contact occurs on the fourth link (or after).  

In the analytical method, as mentioned earlier, we need at least 3 proprioceptive sensors between the contact and the base of the robot in order to be able to recover the force and position values. For the particle filter, there is no such theoretical limitation in the equations and we can apply the method to a contact that occurs on the second link for example. When we do so, however, we see the particles randomly move back and forth along the side of the link where the collision happened, and the force changes with the location of the particle. This comes from the fact that that all of those particle locations and force pairs are equally likely to explain the disturbance torques that we observe and there is not enough information to localize the true contact and force simultaneously.

\subsection{Approach limitations}
It is worth noting here that the particle filter is guaranteed to provide a particle location that is on the link, whereas the analytical solution could return a line of action that does not intersect the robot and, therefore, lead to a non tractable solution if there is too much noise in the model estimates. Additionally, although it is a hard problem and computationally very expensive, the particle filter can be extended to multiple contacts by using several particle sets. 

The main disadvantage of the particle filter is that it is computationally much more expensive and takes longer to converge than the analytical approach. The particle filter requires solving $n$ optimization problems per time step (where $n$ is the number of particles), whereas the analytical approach solves only one optimization per time step. Even in this simple example, we had to run the particle filter slower (100Hz) than the controller (1 kHz) and on a separate thread.

\section{CONCLUSIONS}
This paper shows two different methods to detect, locate, and estimate the force at a contact point in a planar robotic arm. We use an analytical method as a benchmark for a particle based approach. Both approaches use only proprioceptive sensing. 

In simulation, we are able to demonstrate successful localization of a contact point and accurate force estimation for single point collisions during the performance of two different tasks. We showed that both methods work very well, and talked about their advantages and drawbacks. It is worth mentioning that both methods would immediately extend to a non planar robot.

As a future work, it would be interesting to test them on a non planar robot, and on real robots.

We believe that being able to account for collisions with the environment in real time is crucial for the development of robots that can perform complex tasks in unstructured environments.

\addtolength{\textheight}{-12cm}   


\section*{ACKNOWLEDGMENTS}
Thank you to Margot's fairwell mojito for helping boost our creativity.


\bibliographystyle{unsrt}
\bibliography{references}

\end{document}